\UseRawInputEncoding
\documentclass[preprint,12pt]{elsarticle}
\usepackage{graphicx}
\usepackage{amssymb}
\usepackage{epstopdf}
\usepackage{listings}
\usepackage{float}
\usepackage{lineno}
\usepackage{etex}
\usepackage{graphicx,color,colordvi}
\usepackage{latexsym,amsbsy,epsfig,fancybox}
\usepackage{amstext}
\usepackage{subfigure}
\usepackage{amsfonts,textgreek}
\usepackage{mathrsfs}
\usepackage{mathtools}
\usepackage{stmaryrd,dsfont}
\usepackage{bbm}
\usepackage{url}
\usepackage{wrapfig,bm}
\usepackage{tikz}
\usepackage{pstricks}
\usepackage{caption}
\usepackage{psfrag}
\usepackage{upgreek}
\usepackage{isomath}
\usepackage{latexsym}
\usepackage{array}
\usepackage{multirow}
\usepackage{booktabs}
\usepackage{verbatim}
\usepackage{natbib}
\biboptions{sort&compress}
\usepackage{color}
\usepackage{colortbl}
\usepackage[colorlinks=true,citecolor=blue]{hyperref}
\usepackage{overpic}
\usepackage{enumerate}
\usepackage{amsmath,amssymb,xspace}
\usetikzlibrary{plotmarks}
\usetikzlibrary{fit,positioning,arrows}
\usetikzlibrary{shapes.geometric, plotmarks}
\usepackage{appendix}

\usepackage[margin=3cm]{geometry}
\usepackage{soul}

\newcommand{\ac}[1]{\textcolor{red}{add citation}}

\tikzstyle{nicebox}=[draw=black!100, fill=white!10, rectangle, inner sep=4pt, inner ysep=16pt]
\tikzstyle{niceboxtitle}=[draw=black!100, fill=white, text=black, rectangle]

\usetikzlibrary{arrows,decorations.markings}

\newcommand{\figfolder}{RE_figures}
\newcommand{\revision}[1]{#1}

\journal{arXiv}
\begin{document}
\begin{frontmatter}

\title{Generalizability of Graph Neural Network Force Fields for Predicting Solid-State Properties}
\author[1]{Shaswat Mohanty}
\author[1,2]{Yifan Wang}
\author[1]{Wei Cai\corref{cor1}}
	
\address[1]{Department of Mechanical Engineering, Stanford University, CA 94305-4040, USA}
\address[2]{Department of Materials Science and Engineering, Stanford University, CA 94305-4040, USA}

\cortext[cor1]{Corresponding author}

\begin{abstract}
Machine-learned force fields (MLFFs) promise to offer a computationally efficient alternative to {\it ab initio} simulations for complex molecular systems. However, ensuring their generalizability beyond training data is crucial for their wide application in studying solid materials. This work investigates the ability of a graph neural network (GNN)-based MLFF, trained on Lennard-Jones Argon, to describe solid-state phenomena not explicitly included during training. We assess the MLFF's performance in predicting phonon density of states (PDOS) for a perfect face-centered cubic (FCC) crystal structure at both zero and finite temperatures. Additionally, we evaluate vacancy migration rates and energy barriers in an imperfect crystal using direct molecular dynamics (MD) simulations and the string method. Notably, vacancy configurations were absent from the training data. Our results demonstrate the MLFF's capability to capture essential solid-state properties with good agreement to reference data, even for unseen configurations. We further discuss data engineering strategies to enhance the generalizability of MLFFs. The proposed set of benchmark tests and workflow for evaluating MLFF performance in describing perfect and imperfect crystals pave the way for reliable application of MLFFs in studying complex solid-state materials.
\end{abstract}


\end{frontmatter}
\section{Introduction} 
\label{sec:Intro}
Machine learning approaches to aid molecular dynamics (MD) simulations have developed rapidly in recent years. 
Techniques such as deep learning have been applied to accelerated MD simulations~\cite{wang2018deepmd,zheng2021learning,do2022glow,perez2009accelerated} to address their inherent time scale limitations.
Creating accurate force fields has been a key area of development using deep learning architectures, often referred to as machine-learned force fields (MLFFs) or interatomic potentials (MLIPs). 
%
These models aim to achieve the scalability of classical MD simulations with the chemical accuracy of \emph{ab initio} methods. 
\revision{However, verifying that the machine-learned models trained on data from simpler and less expensive empirical potentials can reproduce their behaviors can lead to valuable insights and important guidelines for constructing reliable MLFFs and MLIPs~\cite{mohanty2023evaluating}}.
Graph neural networks (GNNs) are a common architecture choice for MLFFs to encode the local atomic environments \revision{(short and long-range interactions)}.
Starting from SchNet~\cite{schutt2018schnet}, developments along this direction subsequently lead to DeeP-MD~\cite{wang2018deepmd,marcolongo2020simulating,balyakin2020machine}. 
The ability to differentiate between charged and uncharged species, needed for modeling ionic solvents, was introduced by ANI~\cite{smith2017ani,smith2020ani} and further developed into SANI~\cite{stevenson2019schr,jacobson2022transferable} and QRNN~\cite{dajnowicz2022high, mohanty2023development} to capture complex behavior such as polarization and chemical reactivity~\cite{coley2019graph}.
These MLFFs and MLIPs have been used to study a number of systems such as metals~\cite{kruglov2017energy,zeni2019machine,deringer2020general}, biomolecules~\cite{gkeka2020machine}, polymers~\cite{mohanty2023development} and organic molecules~\cite{rosenberger2021modeling}.
Our recent work presented a systematic approach for testing the transferability of these MLFF/MLIP models through the \href{https://gitlab.com/micronano_public/tb-mlff}{TB-MLFF} library~\cite{mohanty2023evaluating}. 
However, existing tests primarily focus on liquid-phase structures and dynamics.
Our prior solid-phase tests were limited to the phonon density of states (PDOS) for a perfect crystal at zero temperature.
Since a significant potential application of MLFF/MLIP models involves finite-temperature modeling of solid materials with defect-controlled properties, more rigorous testing in this domain is essential before trusting these models for computational materials science applications.

This work introduces a more comprehensive set of solid-phase validation tests for MLFFs.
We continue to use the GNN framework developed by \citet{li2022graph} to train an MLFF on data from the Lennard-Jones potential model of Ar, as a representative example. 
We extend our investigation beyond perfect crystals at zero temperature to include lattice vibration at finite temperatures and vacancy migration in imperfect crystals.
On the finite-temperature lattice vibrations, we calculate the phonon dispersion relation, sensitive to both harmonic and anharmonic effects~\cite{debernardi1995anharmonic,turney2009predicting}, using the spectral energy density method (SED)~\cite{thomas2010predicting,ma2018tunable,mcconnell1995mt} implemented in our \href{https://gitlab.com/micronano_public/tbc-xpcs}{C-XPCS} library~\cite{Mohanty2022}. \revision{Similar calculations have also recently been implemented using DeeP-MD simulations~\cite{malgope2024untangling,gupta2023distinct}.}
We also use the spectral energy density to estimate the thermal conductivity of the crystal~\cite{naghdiani2023lattice} and benchmark the performance of the MLFF.
Regarding defect dynamics and the analysis of rare events in imperfect crystals, we calculate the vacancy jump rate from MD simulations using both the MLFF and the original potential model.
The results are also compared against rate theory predictions~~\cite{eyring1935activated} with activation energy barrier computed from a modified version of the string method~\cite{kang2014stress,kuykendall2016investigating,ryu2011entropic,ryu2011predicting,kuykendall2020stress}.
These tests allow us to benchmark the MLFF in the solid phase and can help determine the modifications to the dataset that might be required to make the MLFF more generalizable.

\revision{While in classical force fields, the atomic forces are computed as the negative derivatives of a potential energy function, some of the GNN-based models directly compute the forces without having a potential energy~\cite{li2022graph}.  In other words, the forces by these GNN-based models may be non-conservative, i.e. a potential energy function may not exist.
In this case, one may argue whether such models can be designated as force fields (MLFFs), or perhaps calling them ``force predictors'' is more accurate.
In this paper, we shall still refer to such models as MLFFs, in keeping with the existing literature.
Furthermore, since such models are trained on data obtained from conservative forces, their predicted forces are nearly (although not exactly) conservative, so that the forces should be close to the negative derivatives of some potential energy function.
%
} 

The paper is organized as follows. Section~\ref{sec:Model_form} details the theoretical background of our methods. It describes the methods for computing the Hessian matrix and phonon density of states at zero temperature, as well as the methods for spectral energy density and thermal conductivity prediction at finite temperatures.  It also describes the method for extracting vacancy jump rate from MD simulations, and the approach to estimate it from transition rate theories.
Section~\ref{sec:Numerical} presents the results from our study, including the phonon dispersion curves and the vacancy diffusion rates obtained from both traditional and GNN-based MD simulations.
Section~\ref{sec:conclusion} offers concluding remarks on our findings, and discusses potential future directions for using MLFF to investigate defects in solid materials.
\section{Methods}
\label{sec:Model_form}
\subsection{Phonon density of states} \label{sec:pdos}
The phonon density of states (PDOS) at zero temperature can be obtained from the eigenvalues of the Hessian matrix, $\mathcal{H}$, defined as,
\begin{align}
    \mathcal{H}_{ij}&=\frac{\partial}{\partial x_j} \left(\frac{\partial U}{\partial x_i}\right) 
    = -\frac{\partial {f}_i}{\partial x_j},
    \label{eq:Hij_def}
\end{align}
where $U$ is the potential energy, ${x}_i$ is the coordinate of the $i^{\rm th}$ degree of freedom ($i$ goes from 1 to $3N$ for a system of $N$ atoms), and ${f}_i$ is the atomic force acting on this degree of freedom.
We note that MLFF does not provide the potential energy $U$, but directly predicts the forces $f_i$.
The Hessian matrix can be obtained from numerical differentiation using a centered difference scheme,
\begin{equation}
    \mathcal{H}_{ij} = \frac{{f}_i(x_j-\Delta x)-{f}_i(x_j+\Delta x)}{2 \Delta x},
\end{equation}
In this work, we choose $\Delta x = 3.405 \times 10^{-4}$ \AA \, (i.e. $10^{-4}$ in LJ unit) for the Hessian calculations using LAMMPS.
While too large of a $\Delta x$ can cause a substantial error in the numerical differentiation, care should be taken to ensure that $\Delta x$ is not too small to cause numerical underflow, because MLFF usually uses single-precision floating point numbers.

The Hessian matrix $\mathcal{H}$ as defined in Eq.~(\ref{eq:Hij_def}) is expected to be symmetric ($\mathcal{H}_{ij} = \mathcal{H}_{ji}$) for a conservative force field that is derived from an interatomic potential $U$. 
This symmetry is exactly satisfied in the original Lennard-Jones potential.
However, given that an MLFF does not provide the potential energy function $U$, this symmetry is not guaranteed and is worth checking.
We define a measure for symmetricity, $\mathcal{S}$, as in \cite{mohanty2023evaluating}, for a matrix $A$ as
\begin{equation}
    \mathcal{S}(A) = \frac{\| A_{\rm s} \|_2-\| A_{\rm a} \|_2}{\| A_{\rm s} \|_2+\| A_{\rm a} \|_2},
\end{equation}
where $\| \cdot \|_2$ indicates the Frobenius norm, and  $A_{\rm s}$ and $A_{\rm a}$ are symmetrized and anti-symmetrized matrices defined as,
\begin{align}
    A_{\rm s} &= \frac{A+A^{\rm T}}{2},\\
    A_{\rm a} &= \frac{A-A^{\rm T}}{2}.
\end{align}
The symmetricity $\mathcal{S}$ is bounded by $-1\leq \mathcal{S}(A)\leq 1$, where $\mathcal{S}(A)=1$ for a perfectly symmetric matrix and $\mathcal{S}(A)=-1$ for a perfectly anti-symmetric matrix.  
A good-quality MLFF should have its symmetricity $\mathcal{S}(\mathcal{H})$ as close to $1$ as possible.
We found that in our previous work, the choice of numerical differentiation step size $\Delta x$ was too small ($3.405 \times 10^{-6}$ \AA), which caused the computed Hessian matrix to be excessively asymmetric (see Table 3 of~\cite{mohanty2023evaluating}). 

Given the numerically computed Hessian matrix $\mathcal{H}$, we obtain the eigenvalues $\lambda_n$, $n = 1, \cdots, 3N$, from its symmetrized matrix $\mathcal{H}_{\rm s}$. Here $N$ is the number of atoms in the simulation cell.
The eigenfrequencies $\nu_n$ can be obtained from
\begin{equation}
    \nu_n = \frac{1}{2\pi}\sqrt{\frac{\lambda_n}{m}},
\end{equation}
where $m$ is the atomic mass (assuming all atoms have the same mass).
The distribution of the $\nu_n$ corresponds to the PDOS.
Furthermore, the Hessian matrix can also be used to obtain the phonon dispersion relation of the perfect crystal~\cite{kresse1995ab}.

\subsection{Spectral energy density} \label{sec:SED}
The PDOS and phonon dispersion calculations (based on the Hessian matrix) presented in the previous section rely on the harmonic approximation (HA) of interatomic interactions, which is strictly accurate only in the zero-temperature limit.
Within HA, phonon modes have infinite lifetimes.
However, at finite temperatures, phonon modes exhibit finite lifetimes. The spectral energy density (SED) method~\cite{thomas2010predicting} allows for capturing the phonon dispersion relation at finite temperatures.  
The SED, $\phi(\bm{k},\omega)$, measures the average kinetic energy per primitive cell as a function of wave vector $\bm{k}$, and the frequency $\omega$, where $\omega = 2\pi\nu$.
SED is computed by projecting the atomic trajectories in Molecular Dynamics (MD) simulation of a crystal on the normal vibrational modes of the lattice, as follows. 
\begin{equation}
    \phi(\bm{k},\omega) = \frac{1}{4\pi\tau_0N_{\rm T}}\sum_a
    m
    \bigg\vert \int_0^{\tau_0}\sum_{n_{i,j,k}}^{N_{\rm T}}
    \dot{u}_a(n_{i,j,k};t) 
    \cdot \exp \left[i\bm{k}\cdot
    \bm{r}_0(n_{i,j,k})
    -i\omega t\right] dt \bigg\vert^2, \label{eq:SED_qn}
\end{equation}
where $\tau_0$ is the total simulation time, $m$ is the atomic mass, and $N_{\rm T}$ is the total number of primitive cells in the simulation cell.  $N_{\rm T}=N_x\times N_y\times N_z$, where $N_x$, $N_y$, and $N_z$ refer to the number of primitive cells in the $x$, $y$, and $z$ directions, respectively. $n_{i,j,k}$ denotes primitive cells with $i\in[1,N_x]$, $j\in[1,N_y]$, and $k\in[1,N_z]$. 
Eq.~(\ref{eq:SED_qn}) is only applicable to crystals with a single-atom basis, such as the face-centered cubic (FCC) crystal of Ar considered here.
$\dot{u}_a$ stands for the atomic velocity in the $a$-th direction, where $a = x$, $y$, or $z$. $\bm{r}_0(n_{i,j,k})$ denotes the equilibrium position of the $n_{i,j,k}$-th primitive cell. 
The resulting $\phi(\bm{k},\omega)$ map exhibits peaks at the ($\bm{k}$, $\omega$) loci that correspond to the phonon dispersion relation. 

To ensure accurate SED calculations using Eq.(\ref{eq:SED_qn}), the MD trajectory should cover a time period longer than the lifetime of the phonon mode of interest. For a phonon mode with wave vector $\bm{k}_0$, the SED profile's peak (with respect to $\omega$) can be fitted with a Lorentzian function to determine the phonon mode's lifetime,
 \begin{equation}
     \phi(\bm{k}_0,\omega) =  \frac{\phi_{\rm max}}{1+\left(\frac{\omega-\omega_c}{\gamma}\right)^2}, \label{eq:phon_life}
 \end{equation}
where $\phi_{\rm max}$ is the peak value, $\omega_c$ is the peak frequency, and $\gamma$ is the half-width at half maximum. From the fit, the phonon mode's lifetime can be estimated as $\tau=\frac{1}{2\gamma}$.
The SED method serves as a valuable tool for assessing the accuracy of a force field in capturing lattice vibration statistics and reproducing the phonon dispersion relation and phonon lifetimes at finite temperatures.

Given the SED data, we can further predict the lattice thermal conductivity, $\kappa$, using the following expression~\cite{naghdiani2023lattice}.
\begin{equation}
    \kappa = \sum_{(\bm{k},\,\omega)} C_v(\omega) \, v_g^2(\bm{k},\omega)\, \tau(\bm{k},\omega). \label{eq:kappa}
\end{equation}
where $C_v=\frac{k_{\rm B}x^2 e^x}{V[e^x-1]}$, $x = \frac{\hbar \omega}{k_{\rm B}T}$ and $V$ is the volume of the simulation cell. The sum is carried out over all $3N$ pairs of $(\bm{k},\omega)$ that satisfy the phonon dispersion relation.
$\hbar$ is the reduced Plank's constant, $k_{\rm B}$ is the Boltzmann constant, and $\tau$ is the lifetime of the phonon in mode $(\bm{k}, \omega)$. $v_g$ is the group velocity which is obtained by evaluating the following derivative numerically along the phonon dispersion relation, $\omega(\bm{k})$.
\begin{equation}
    v_g = \frac{d \omega}{d k} 
\end{equation}
where $k=\|\bm{k}\|$.
Given the dependence of $\kappa$ on the lifetime of the phonon in all the vibrational modes, the accurate estimation of $\kappa$ serves as a rigorous test for an MLFF's ability to capture lattice dynamics.

\subsection{Vacancy Jump Rates} 
\label{subsec:rate_theory}
We will use the tests described in the proceeding sections to measure the MLFF's ability to capture the lattice dynamics of a perfect crystal.
To assess its performance for a defective crystal, we will focus on estimating the migration rate of a single vacancy. 

To determine the vacancy migration rate, we conducted a long MD simulation ($\sim53.9$ ns) of a crystal containing a vacancy, tracking the vacancy's location over time.
We then sample 100 short trajectories each with a starting time randomly chosen from the simulation.
Defining $\chi(t)$ as the fraction of the vacancies that have not made a jump by time $t$ in these sampled trajectories, we expect $\chi(t)$ to decay exponentially with time.  By fitting $\chi(t) = \exp (-t/\tau_{\rm b})$, we obtain the vacancy jump rate, $r_{\rm j} \approx 1/\tau_{\rm b}$.

After identifying the vacancy jump rate at a given temperature $T$, we repeat the calculation at different temperatures. The resulting data are fitted to an Arrhenius expression,
\begin{equation} 
    r_{\rm j} = \nu_0 \,
    \exp\left(-\frac{E_{\rm b}}{k_{\rm B}T}\right). \label{eq:arrhenius}
\end{equation}
where $\nu_0$ is the rate prefactor and $E_{\rm b}$ is the energy barrier. 
There are various ways to estimate the rate prefactor, such as the Eyring-Polanyi equation~\cite{eyring1935activated} and the harmonic transition state theory~\cite{vineyard1957frequency}.
A rough estimate for the rate prefactor is the Debye frequency $\nu_{\rm D}$, which is related to temperature $\Theta_{\rm D}$ by $\nu_{\rm D} = k_{\rm B} \Theta_{\rm D} / h$, where $h=2\pi\hbar$ is Planck's constant.
Given that the Debye temperature for solid Argon is $\Theta_{\rm D} = 92$~K~\cite{stewart1983measurement}, the Debye frequency is $\nu_{\rm D} = 1.92\times10^{12}$~s$^{-1}$.
The energy barrier is the energy increase from the initial state (A) to the transition state (S), which can be separately predicted by minimum energy path (MEP) searches, such as the string method~\cite{kang2014stress,kuykendall2016investigating,ryu2011entropic,ryu2011predicting,kuykendall2020stress}.
These estimates of $\nu_0$ and $E_{\rm b}$ serve as additional benchmarks to the MLFF's ability to accurately capture vacancy jump rates.

\section{Results}  \label{sec:Numerical}
We present comparisons between the GNN-based MLFF and the original Lennard-Jones (LJ) potential on simulations of solid Argon, which forms a face-centered cubic (FCC) lattice at low temperatures.
The GNN is trained on the LJ force field using samples containing both solid and liquid configurations (from 10~K to 105~K), similar to the previous models~\cite{Mohanty2022}. 
The solid configurations used in the training of GNN-based MLFF correspond to perfect crystals (i.e. vacancy-free) unless mentioned otherwise explicitly. \revision{The MD simulations are performed in non-dimensional or LJ units where the units in distance, time, and energy are given by: $1\sigma=3.4$~\AA, $1\tau=2.156$ ps, and $1\epsilon = 0.010327$ eV, respectively.}

\subsection{Phonon dispersion relations}
\begin{figure}[H]
    \centering
    \includegraphics[width=0.45\textwidth]{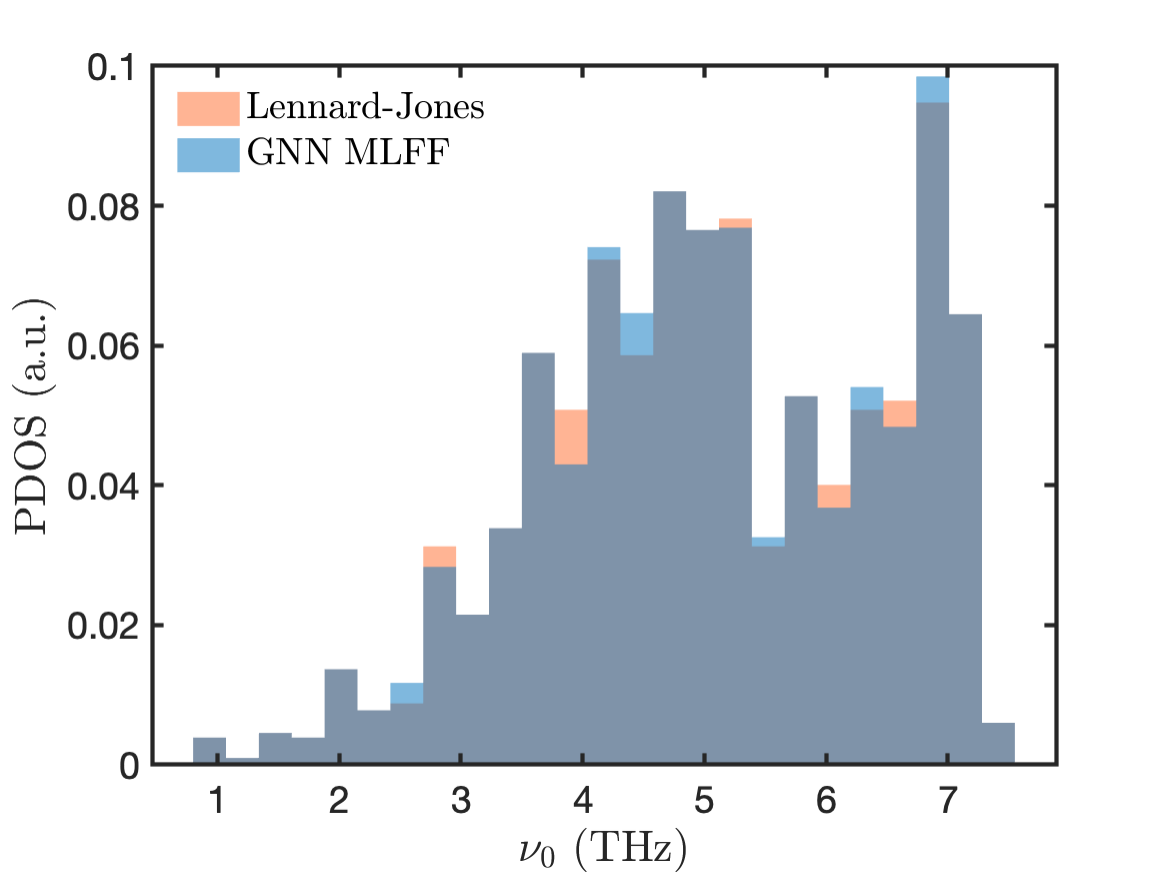}
    \caption{The phonon density of states at $T = 0$~K obtained from Lennard-Jones force field and the GNN-based force field with MAE of 0.02 THz and MaxAE of 0.09 THz.}
    \label{fig:ref_pdos}
\end{figure}

\begin{figure}[H]
    \centering
    \subfigure[]{\includegraphics[width=0.48\textwidth]{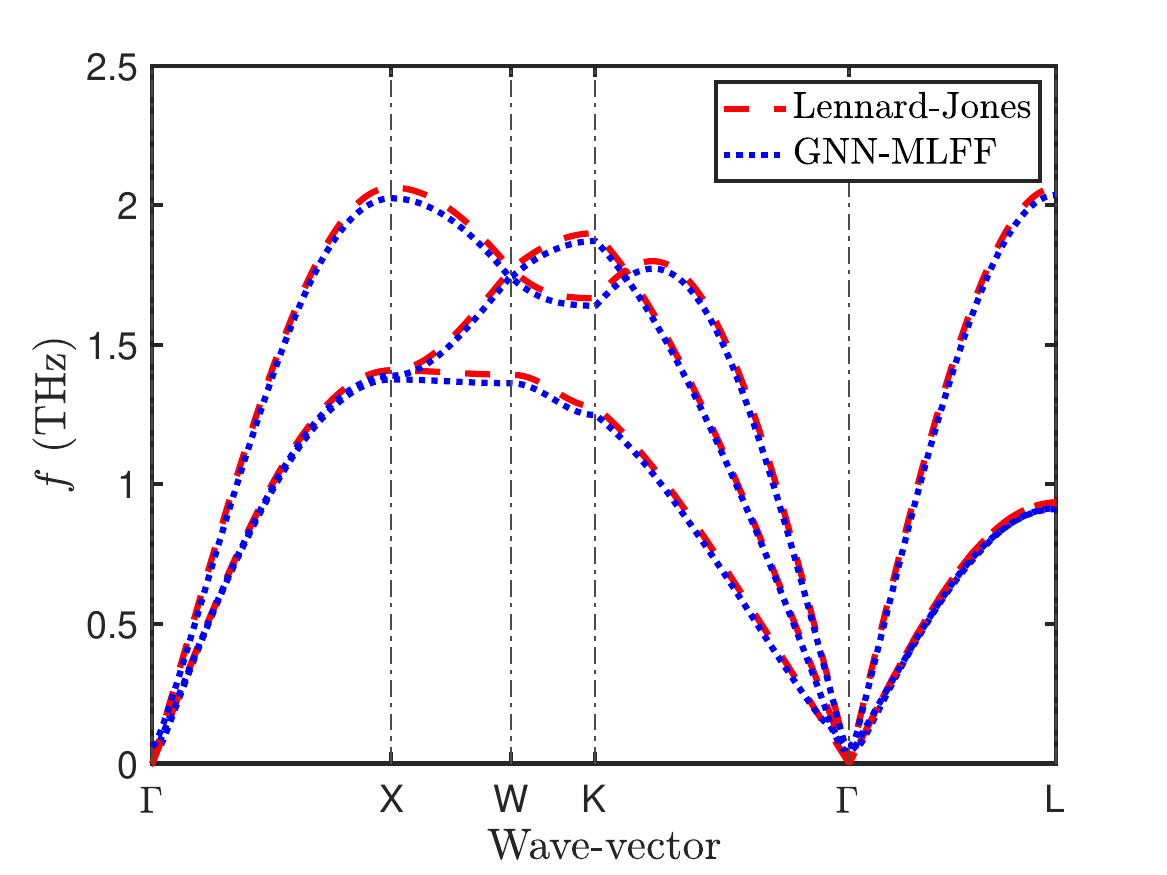}
   \label{fig:pdisp_press}}
   \subfigure[]{\includegraphics[width=0.48\textwidth]{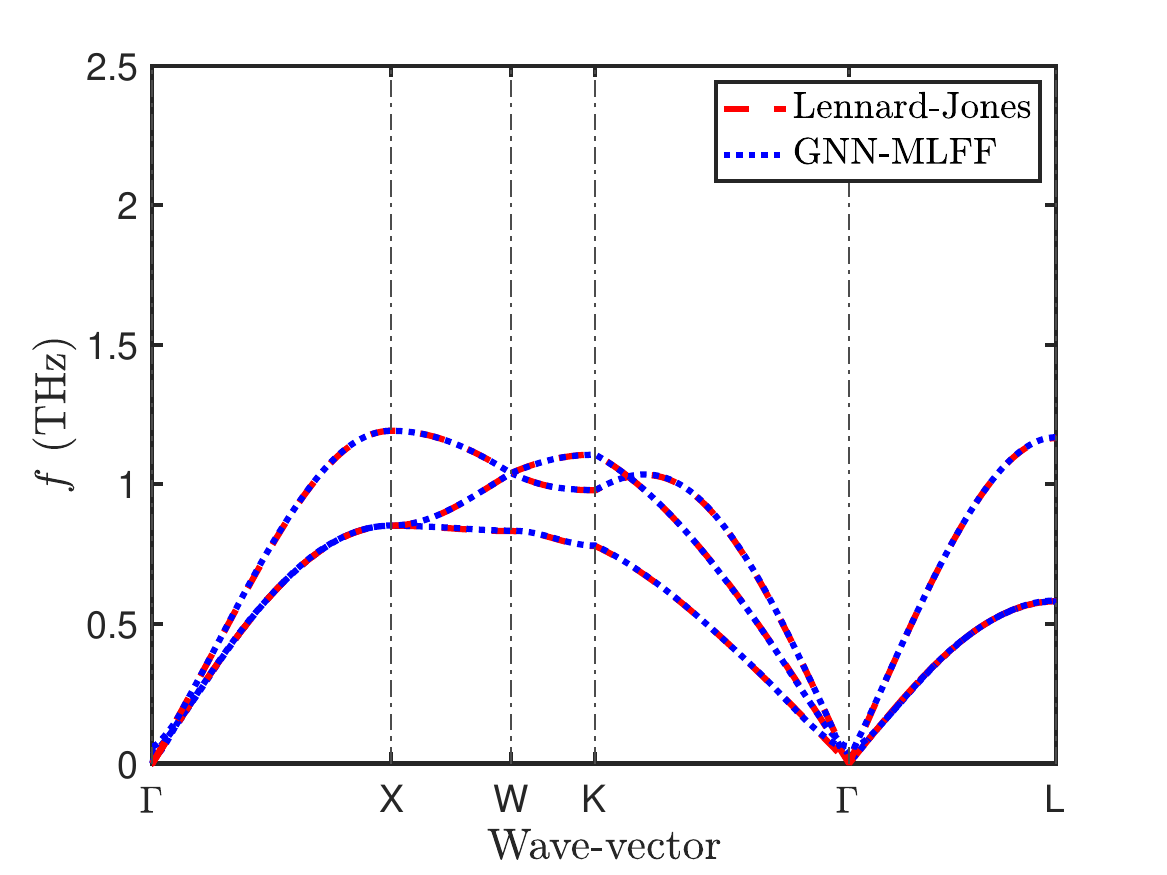}
   \label{fig:pdisp_vac}}
    \caption{The phonon dispersion relation at $T = 0$~K obtained from the reference MD potential and the GNN-based MLFF at (a) $p = 137.1$~bar (training configuration pressure) with MAE of 0.02 THz and MaxAE of 0.07 THz and (b) $p = -2.5$ kbar with MAE of 0.002 THz and MaxAE of 0.06 THz.\label{fig:pdisp}}
\end{figure}

\begin{figure}[H]
    \centering
    \subfigure[]{\includegraphics[width=0.48\textwidth]{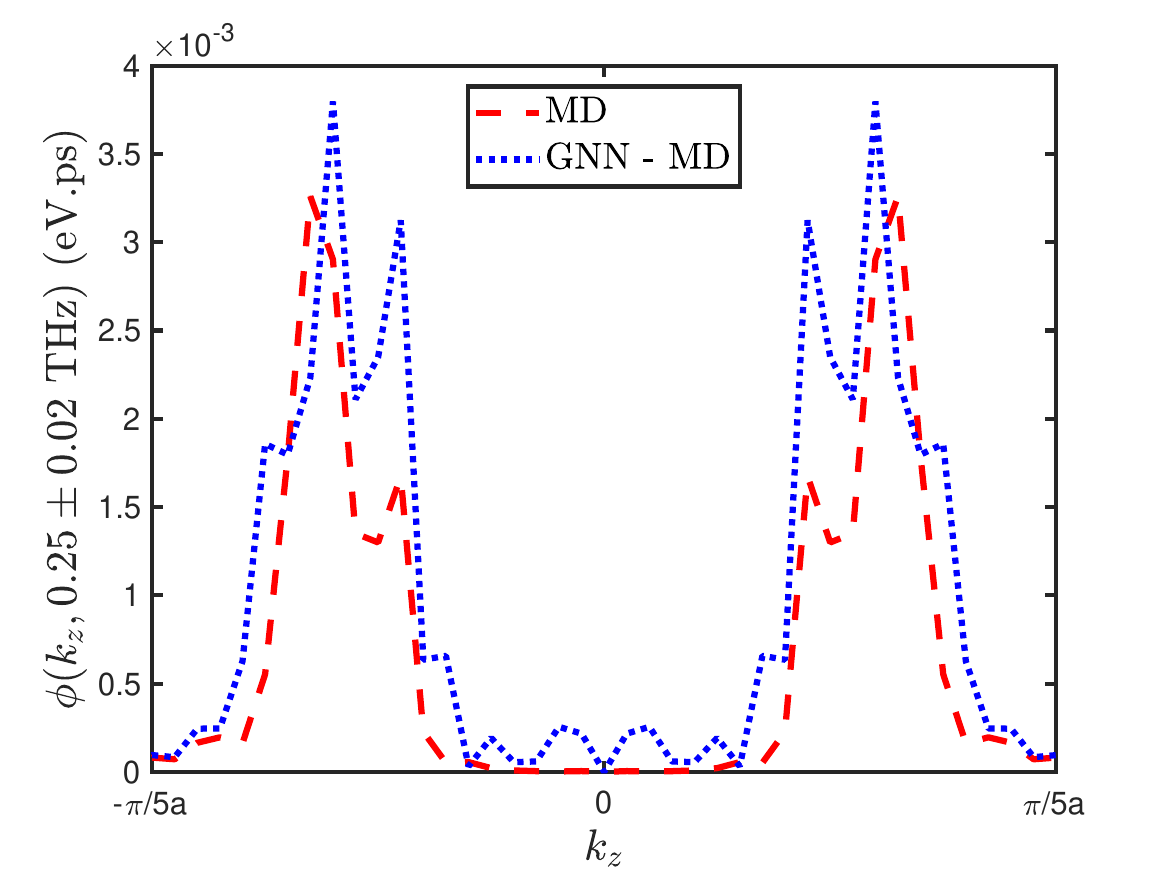}
   \label{fig:phi_k}}
   \subfigure[]{\includegraphics[width=0.48\textwidth]{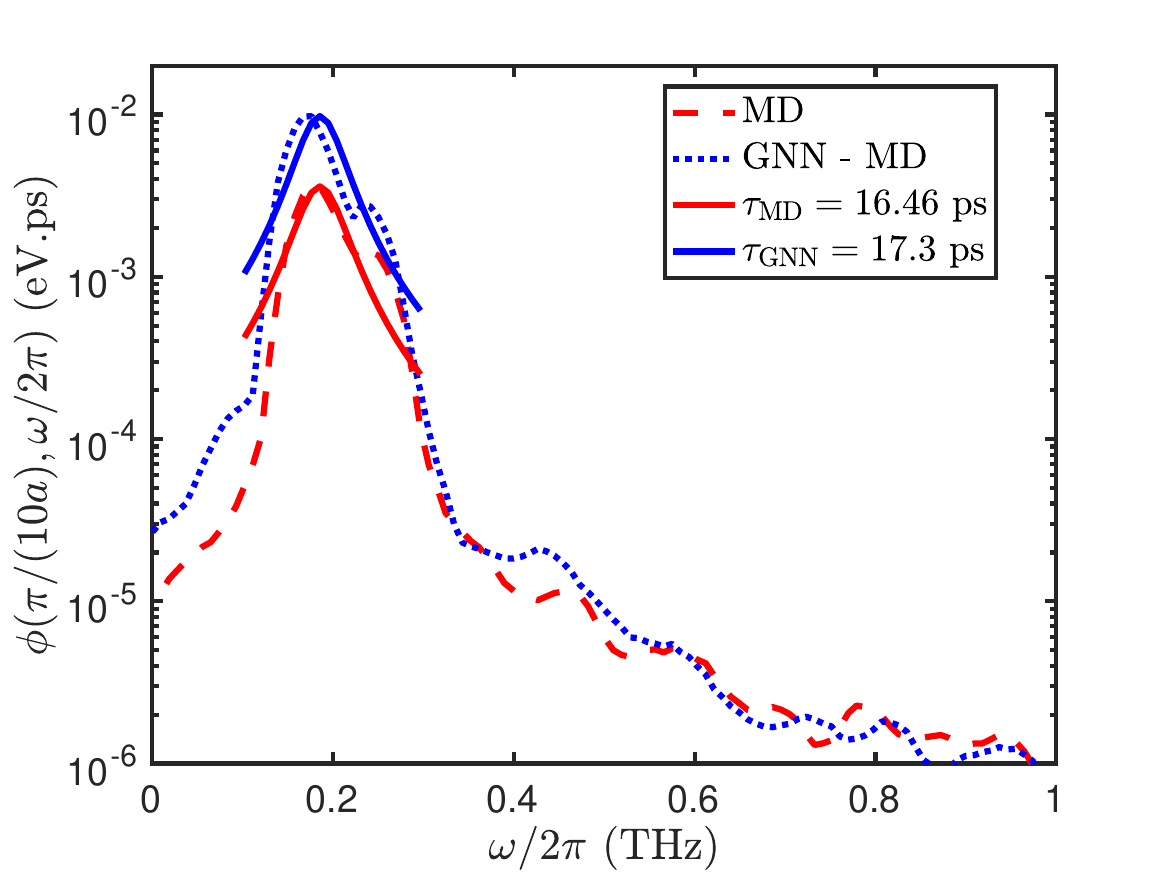}
   \label{fig:phi_w}}
    \caption{The spectral energy density, $\phi(\bm{k},\omega)$, in eV$\cdot$ps, at $T = 40$~K and $p = 137.1$ bar, (a) averaged over the frequency range of $\omega/2\pi = 0.25 \pm 0.02$ THz and (b) calculated at $k_z=\pi/(10a)$.\label{fig:phi_line}}
\end{figure}

As discussed earlier in Secton~\ref{sec:pdos}, the Hessian matrix method can be used to obtain the phonon density of states at 0 K. Using a finite difference step size of $\Delta x = 3.4\times 10^{-4}\,$ \AA~ \revision{($10^{-4}\sigma$)} for the GNN-based MLFF, we obtain a nearly symmetric $\mathcal{H}$ (the deviation of symmetricity $\mathcal{S}$ from $1$ is on the order of $10^{-5}$).
\revision{This result indicates that the forces predicted by the MLFF are nearly (although not exactly) conservative.}
The coarser finite difference step size $\Delta x$ doesn't affect the agreement of the PDOS from the GNN-based MLFF against the reference LJ potential, as shown in Fig.~\ref{fig:ref_pdos}.  The mean absolute error (MAE) for each individual eigenfrequency (for the same eigenmode) is under 0.03 THz and the maximum absolute error (MaxAE) is less than 0.1 THz.
We can obtain the $\mathcal{H}$ and the subsequent PDOS at different pressures by imposing different pressures on the simulation cell. To test the accuracy and the generalizability of the $\mathcal{H}$ obtained from the GNN-based MLFF, we compute the phonon dispersion relation at 137.1 bar (the pressure of the simulation for the training configurations) and at -2.5 kbar, which represents unseen data for the model, as shown in Fig.~\ref{fig:pdisp}. Under both pressures, the MAE and the MaxAE between each eigenfrequency (for the same eigenmode) are less than 0.1 THz.
The Hessian matrix allows us to calculate the phonon dispersion at 0 K where the phonons are trapped in a given mode of vibration (infinite lifetime).
To obtain the phonon dispersion relation at a finite temperature where the phonons have a finite lifetime in any particular mode, we use the spectral energy density (SED) method (see Section~\ref{sec:SED}).
To this end, we consider a supercell that contains $5\times5\times100$ FCC unit cells, with a lattice parameter, $a=4.808$ \AA \revision{($\sqrt{2}\sigma$)} . The simulation cell is extended significantly in the $z$-direction to achieve a fine resolution in $k_z$. This quasi-1D configuration allows us to compute the phonon dispersion as a function of $k_z\in[-\pi/a,\pi/a)$ (from $\Gamma$ point to X point in the reciprocal lattice). The initial configuration is relaxed to an energy minimum configuration by using the conjugate gradient algorithm, to obtain the equilibrium position, $\bm{r}(n_{x,y,z},0)$, for each primitive cell. We use the equilibrium position calculated by the LJ force field for the SED calculations for both the MD simulation and GNN-MD simulation trajectory. We first perform an NVT simulation thermostatted at 40 K (finite temperature) and adjust the simulation cell dimensions every 10 ps to reach the target pressure of $p = 137.1$~bar. We then obtain the equilibrium lattice constant of the crystal at finite temperature by performing energy relaxation. We finally perform an NVT simulation at the equilibrium box size for 100 ps, thermostatted at $T = 40$~K. The equilibrated configuration is used as the initial configuration for the MD and GNN-MD simulations, each carried out under an NVT ensemble thermostatted at 40 K for 107.8 ps \revision{($50 \tau$)}, with configuration snapshots being recorded at 107.8 fs \revision{($0.05 \tau$)} intervals. We use these 1000 frames to compute the phonon dispersion relation using the SED analysis, which is sufficient since it is greater than the phonon lifetimes of solid Argon at this temperature. 
\begin{figure}[H]
    \centering
    \subfigure[]{\includegraphics[width=0.48\textwidth]{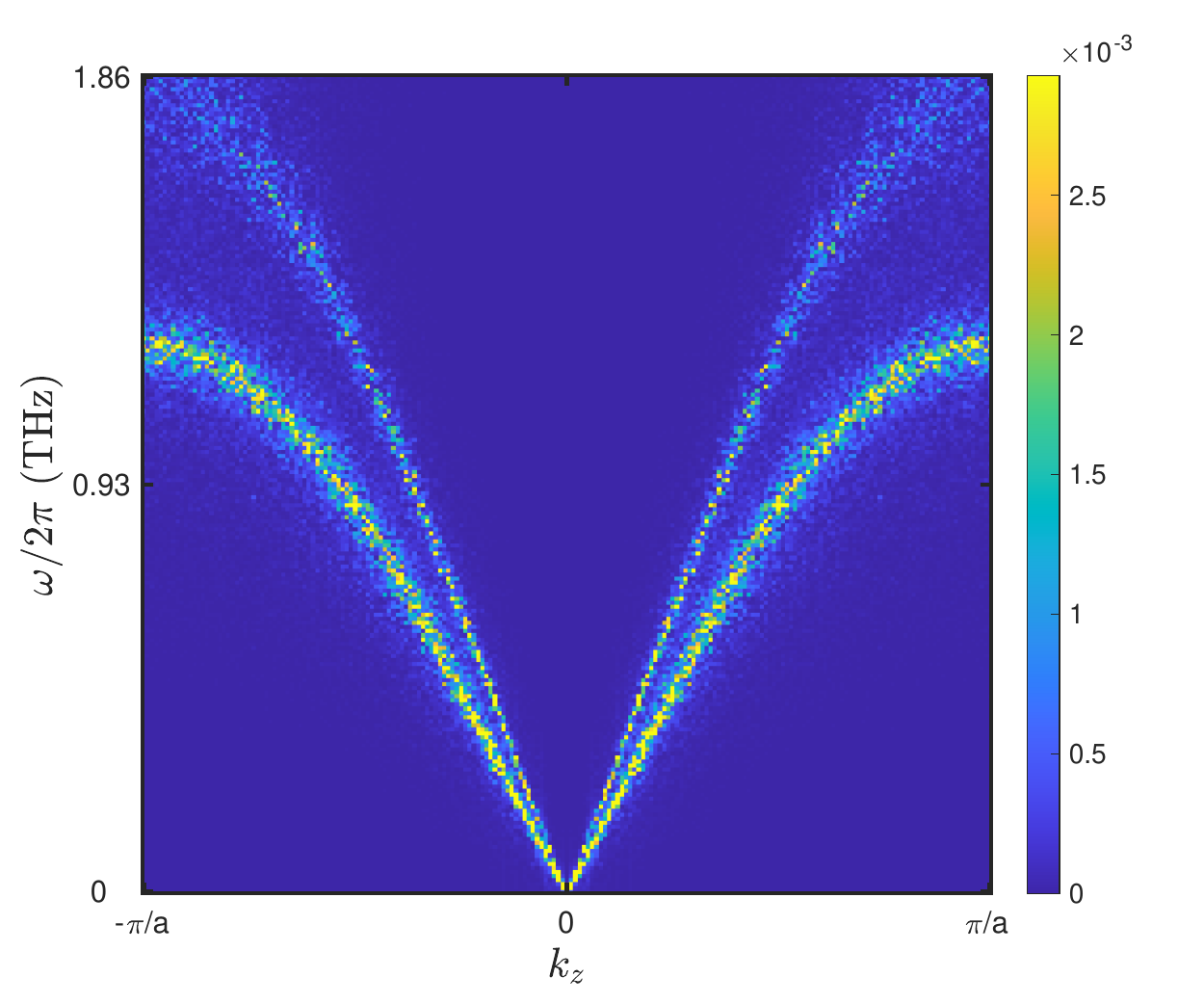}
   \label{fig:phi_md}}
   \subfigure[]{\includegraphics[width=0.48\textwidth]{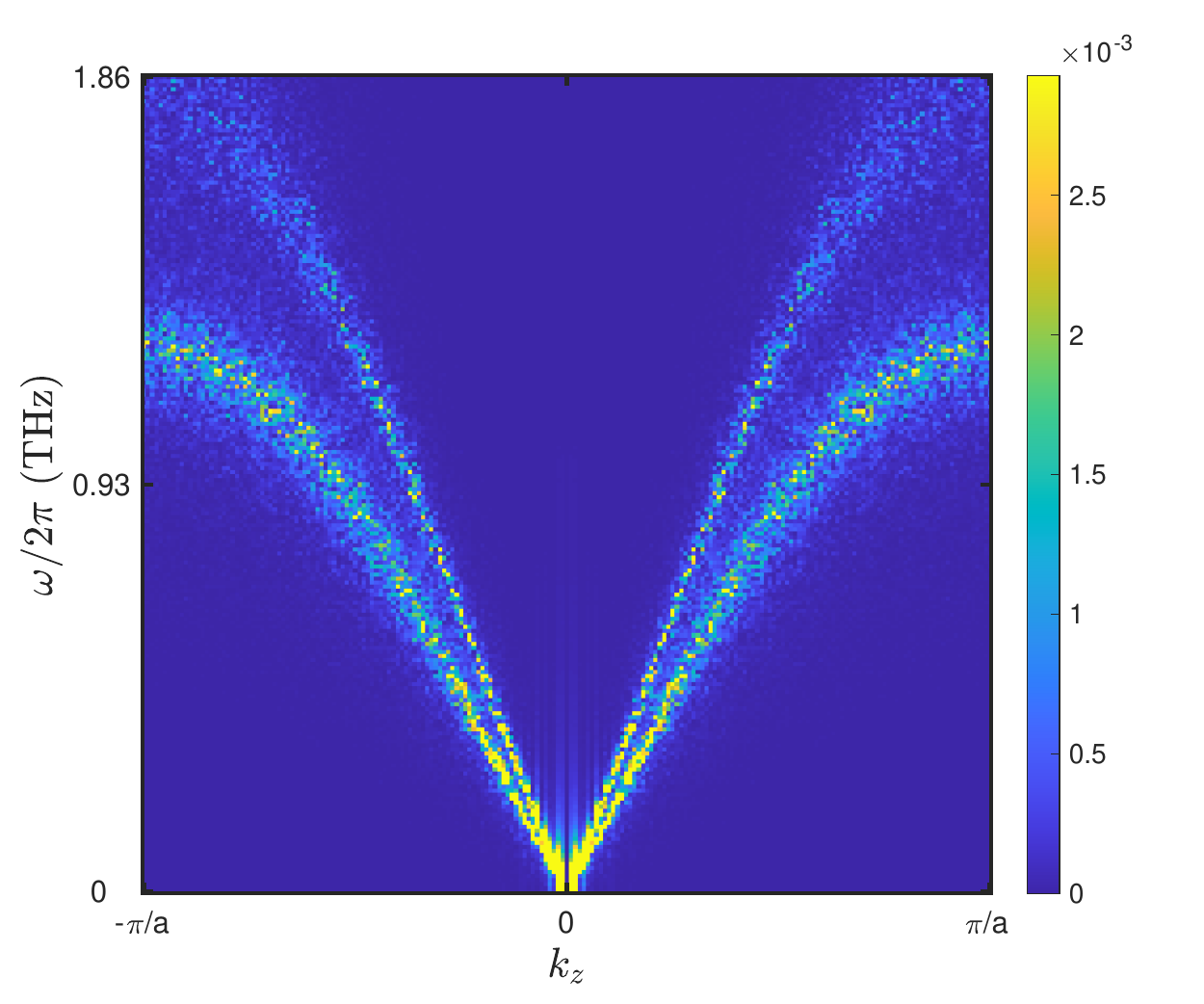}
   \label{fig:phi_gnn}}
    \caption{The spectral energy density, $\phi(\bm{k},\omega)$, in eV$\cdot$ps, 
    at $T = 40$~K and $p = 137.1$ bar, obtained from the (a) MD simulation and (b) GNN-MD simulation trajectories.
    \label{fig:phi}}
\end{figure}
The SED calculated from the MD and GNN-MD trajectory are in close agreement, as shown in Fig.~\ref{fig:phi}, and the phonon dispersion relation obtained from the force constant matrix at 137.1 bar, indicating that the GNN-MD simulation is capable of capturing the acoustic and thermal properties of the system it is trained. The GNN-MD has some artificial oscillations in the SED only around small $k_z$ values. In addition, we also compare the SED at $k_z=\pi/(10a)$ for frequencies under 1 THz and the averaged SED over a frequency range of $0.25\pm 0.02$ THz for $k_z \in [-\pi/a,\pi/a)$, as shown in Fig.~\ref{fig:phi_line}. The peaks in Fig.~\ref{fig:phi_k} correspond to the momentum of the phonon with an angular frequency of $0.25$ THz. On the other hand, the inverse of the fitted peak widths in Fig.~\ref{fig:phi_w} corresponds to the phonon lifetimes, as described in and below Eq.~(\ref{eq:phon_life}). The phonon lifetime corresponding to the phonon mode with a momentum of $\hbar k_z$ obtained from the GNN-MD simulation is within $5\%$ of the lifetime obtained from the MD simulation. 
\subsection{Thermal conductivity predictions}
The SED signal generated from the MD and GNN-MD trajectory can be used to extract the thermal conductivity, $\kappa$, of the system using the relation in Eq.~(\ref{eq:kappa}). We use a cubic simulation cell for the $\kappa$ calculations, where the supercell is made of $10\times10\times10$ FCC unit cells and is thermostatted at 40 K and barostatted at a pressure of 137.1 bar. We limit the size of the simulation cell in the interest of the computational expense of the GNN-MD simulation. However, our results confirm that the simulation cell is large enough for a reliable estimate of $\kappa$. To validate the $\kappa$ obtained from the SED calculations, we compare it against reference calculations from an MD simulation of a larger simulation cell, using the Green-Kubo formulation~\cite{zwanzig1965time}. The reference simulation cell consists of $50\times50\times50$ FCC unit cells thermostatted at 40 K and barostatted at a pressure of 137.1 bar and the resulting thermal conductivity is $\kappa \sim 0.49$ W$\,$m$^{-1}$K$^{-1}$. 
From the SED calculations, we obtain $\kappa_{\rm MD}\sim 0.45$ W$\,$m$^{-1}$K$^{-1}$ from the MD trajectory and $\kappa_\textrm{GNN-MD}\sim 0.55$ W$\,$m$^{-1}$K$^{-1}$ from the GNN-MD trajectory, which are both within $15\%$ of the reference calculations. \revision{Note that the thermal conductivity calculations are carried over all the $\phi(\bm{k},\omega)$ pairs on the 3D Fourier grid (wave-vector space) and are not just restricted to the high symmetry directions.}
Overall, the phonon dispersion and thermal conductivity calculations from the MD and GNN-MD trajectories are in good agreement.
These results provide an estimate of the quality of the GNN-based MLFF in thermal conductivity predictions and illustrate the value of these benchmarking tests for evaluating solid-state behavior in MLFF or MLIP models.

\subsection{Vacancy jump rates}
Having tested the performance of the MLFF for perfect crystals, we now 
present the analysis of the vacancy jumping rate in an imperfect crystal (containing a single vacancy). 
We started with a 256-atom solid configuration (in FCC lattice) and then removed one atom to create a vacancy.
We also imposed a negative pressure of $\sim 2.5$ kbar, resulting in  a low-density simulation cell at $5.711$ mol/cm$^3$.
The low-density simulation cell is chosen to reduce the energy barrier for vacancy jumps, so that they can be observed in the limited timescales accessible by MD and GNN-MD simulations. 
We use the resulting configuration to run MD and GNN-MD simulations for $\sim 53.9$ ns \revision{($2.5\times10^4\, \tau$)} where we record the configuration snapshot every $\sim 1.078$ ps \revision{($0.5\, \tau$)}. 
We employ the Wigner-Seitz analysis using OVITO~\cite{stukowski2009visualization} to identify the vacancy in each configuration by comparing it to a perfect crystal configuration. 
A vacancy jump is recorded whenever the atom ID of the zero-occupancy atom changes in the Wigner-Seitz analysis.

\begin{figure}[H]
    \centering
    \subfigure[]{\includegraphics[width=0.48\textwidth]{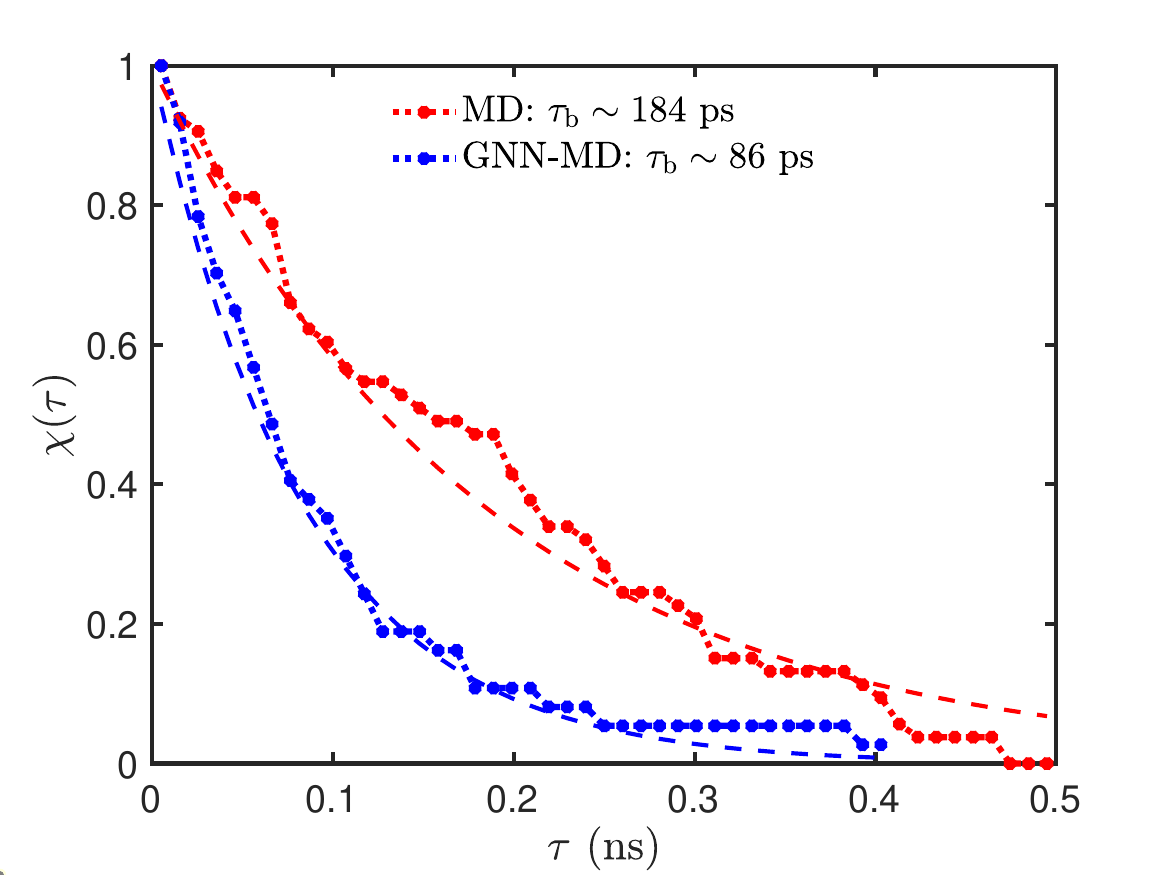}
   \label{fig:unbond_frac}}
   \subfigure[]{\includegraphics[width=0.48\textwidth]{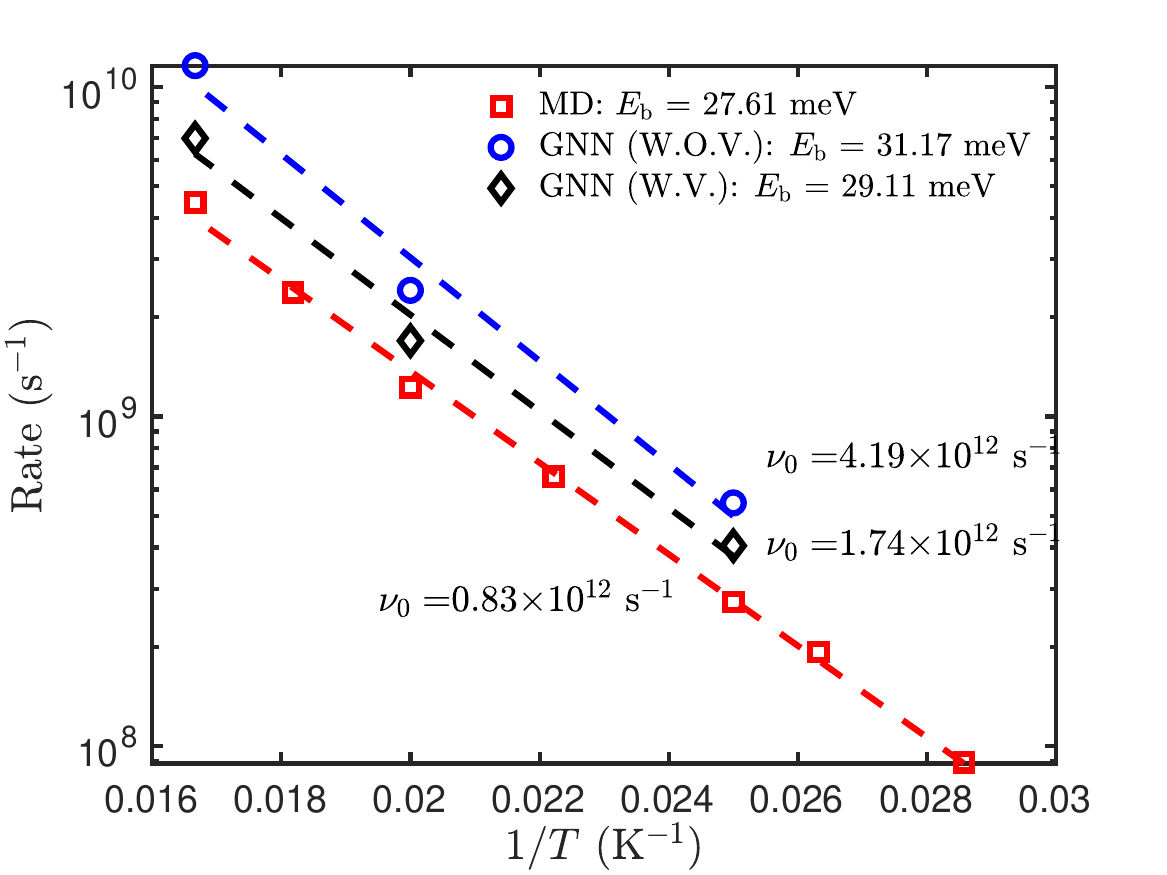}
   \label{fig:rate_fig}}
    \caption{(a) The fraction of simulation trajectories in which the vacancy has not jumped as a function of time at $T = 60$~K from the GNN-MD (W.O.V.) simulation. (b) The Arrhenius plot of vacancy jumping rate as a function of $1/T$ .
    GNN-MD (W.O.V.) and GNN-MD (W.V.) correspond to the MD simulation using the model trained on configurations without and with a single vacancy, respectively.
    \label{fig:vac_diff}}
\end{figure}

We calculate the vacancy jump rate as described in Section~\ref{subsec:rate_theory}.
Fig.~\ref{fig:unbond_frac} shows the fraction $\chi$ of independent simulations in which the vacancy has not jumped as a function of time $\tau$.
The mean time for vacancy jumping, $\tau_{\rm b}$, is obtained by fitting $\chi(\tau)$ to $ \exp (-\tau/\tau_{\rm b})$.
We note that under this condition ($T = 60$~K) the mean jump time in GNN-MD is about a factor of 2 shorter than the original MD model, meaning that the vacancy jump rate predicted by the GNN-MD is about a factor of 2 greater than the original MD model.
Nonetheless, that the GNN-MD predicts the vacancy jump rate within the same order of magnitude as that from the original MD model is an encouraging result, because it is common for rates to vary by orders of magnitude.
We repeat the calculation at different temperatures and extract the vacancy jump rate.  Fig.~\ref{fig:rate_fig} shows the Arrhenius plot of the rate as a function of inverse temperature predicted by the MD and GNN-MD simulation.
Since the MD simulations are relatively inexpensive, they are carried out at 35 K, 38 K, 40 K, 45 K, 50 K, 55 K, and 60 K. On the other hand, the GNN-MD simulations are restricted to 40 K, 50 K, and 60 K. 
The predictions for each model can be well fitted by the Arrhenius expression, $r_{\rm j} = \nu_0 \exp (-{E_{\rm b}}/({k_BT}))$, where $\nu_0$ is the frequency prefactor and $E_{\rm b}$ is the energy barrier. 
The curve labelled as GNN-MD (W.O.V.) corresponds to the GNN-based MLFF trained on solid and liquid configurations in which the solid is a perfect crystal that does not contain any vacancies. This is the same model used in all the previous plots, including Fig.~\ref{fig:unbond_frac}.
The curve labelled as GNN-MD (W.V.) corresponds to a GNN-based MLFF trained on solid and liquid configurations in which $15 \%$ of the dataset correspond to a crystal containing a single vacancy.

Our results show that both GNN-based models (with or without vacancy in the training data) are within 4 meV of the $E_{\rm b}$ predicted by the Arrhenius fit on the MD data of the original LJ potential.  
The frequency prefactor $\nu_0$ computed from the GNN-MD simulations is within one order of the MD simulation results using the original LJ potential.
The latter is also within one order of magnitude of the rough estimate provided by the Debye frequency $\nu_{\rm D}$.
The GNN-MD simulations somewhat overpredict the vacancy jumping rate (hence $\nu_0$) compared to the reference MD calculations. 
The inclusion of the vacancy configurations in the training data helps reducing the mismatch in the vacancy jump rates, as shown in Fig.~\ref{fig:rate_fig}. 
It is remarkable that the GNN-MD (W.O.V.) model, which did not see any vacancy configurations in a crystal during its training, performs reasonably well in terms of predicting the vacancy jump rates.
Furthermore, it is unlikely that during training the GNN-MD (W.V.) model has seen any configurations corresponding to saddle configurations for vacancy jump (because of its rarity).
Given that the activation energy of saddle configurations controls the vacancy jump rate (see Section~\ref{sec:Eb}), the improved agreement in the prediction of vacancy jump rate by the GNN-MD (W.V.) model with respect to the original MD model is also encouraging.

\subsection{Energy barriers for vacancy jump}
\label{sec:Eb}
\begin{figure}[H]
    \centering
    \subfigure[]{\includegraphics[width=0.48\textwidth]{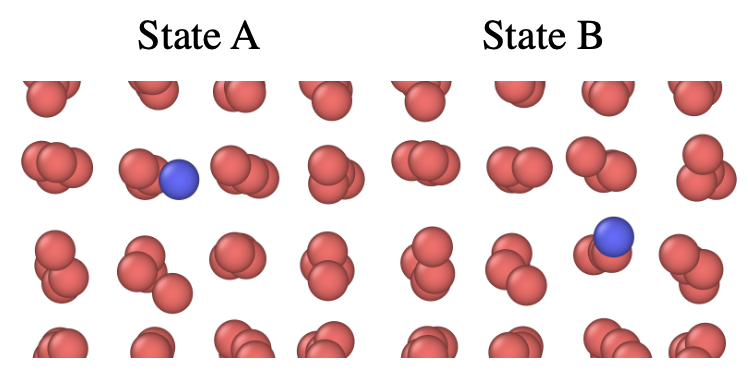}
   \label{fig:vac}}
   \subfigure[]{\includegraphics[width=0.48\textwidth]{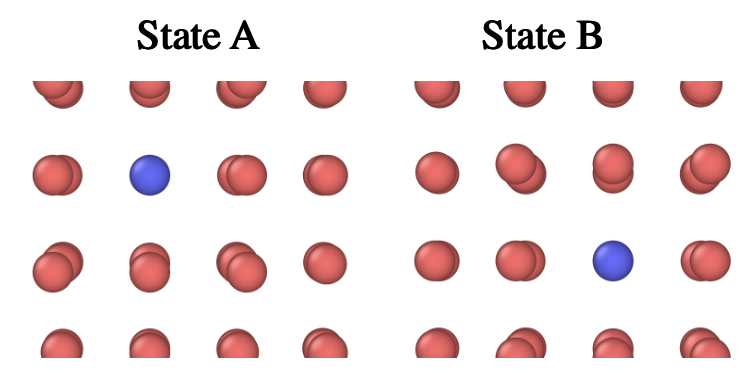}
   \label{fig:MEP}}
    \caption{Initial and final configurations illustrating the vacancy jump (highlighted by the virtual blue atom). (a) Snapshots from MD simulations at $T = 40$~K. (b) Energy-minimized configurations used by the string relax method.  
    \label{fig:vac_mep}}
\end{figure}

\begin{figure}[H]
    \centering
    \subfigure[]{\includegraphics[width=0.48\textwidth]{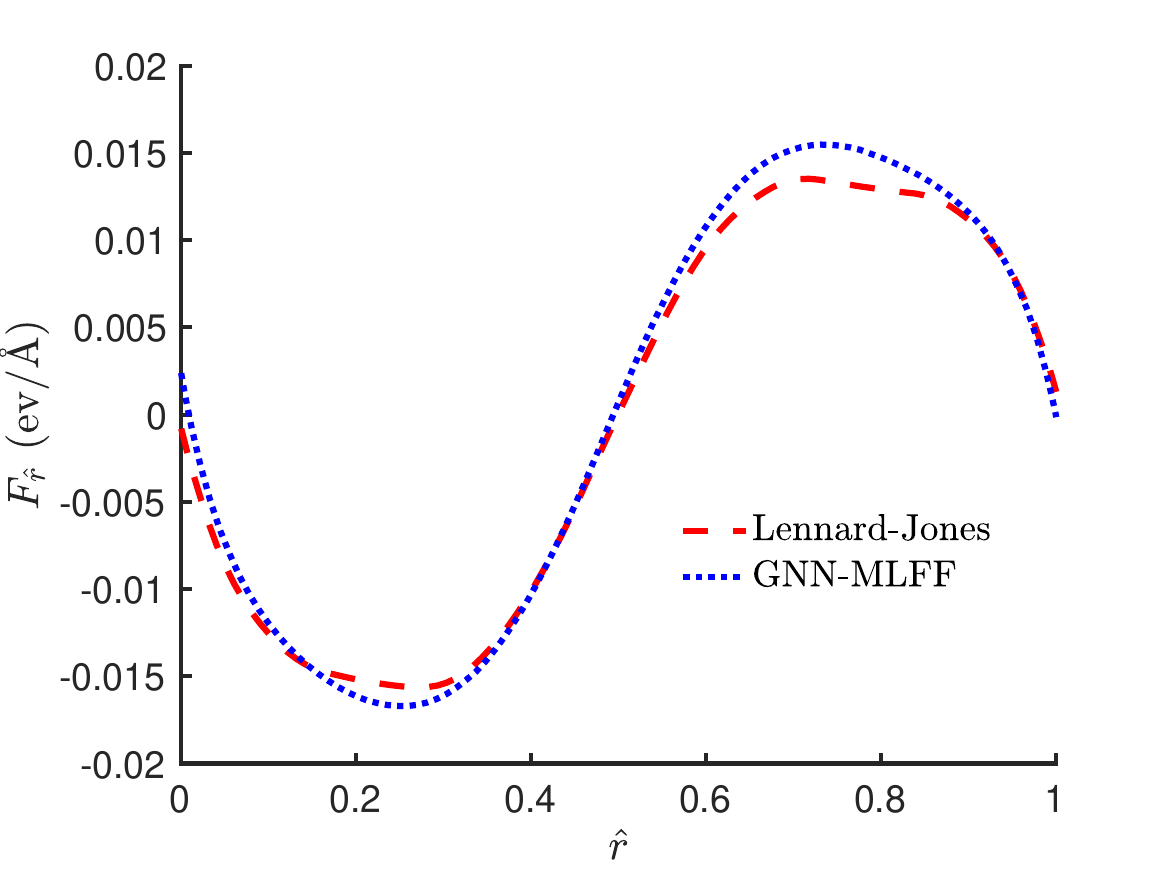}
   \label{fig:MEP_force}}
   \subfigure[]{\includegraphics[width=0.48\textwidth]{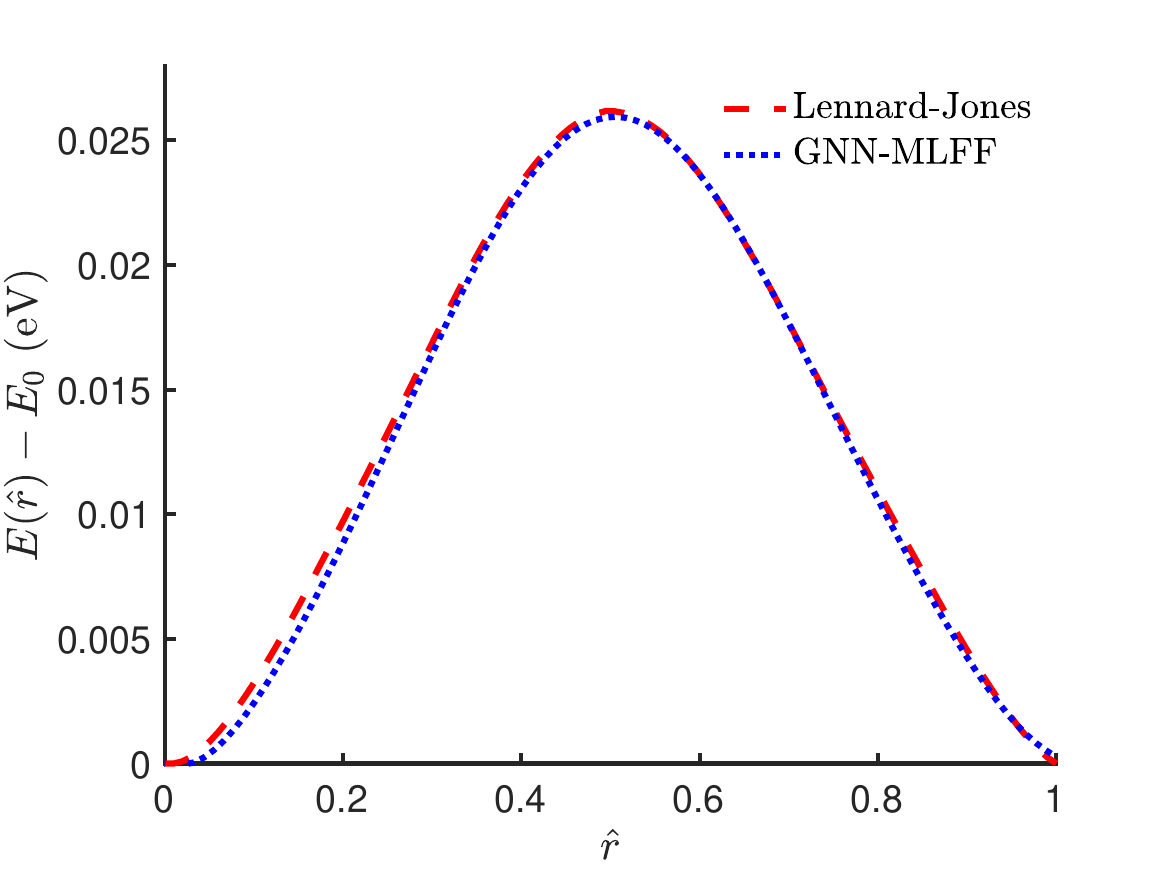}
   \label{fig:MEP_energy}}
    \caption{(a) Force along the MEP calculated by the LJ and GNN force calculators. (b) The energy barrier constructed by numerically integrating the force along the MEP, where $E_{\rm b} = 26.18$~meV for the original LJ potential, and $E_{\rm b} = 25.95$~meV for the GNN-based (W.O.V.) MLFF. \label{fig:MEPS}}
\end{figure}

To benchmark the rates obtained from MD simulations against theories, we separately compute the energy barrier $E_{\rm b}$ by searching for the minimum energy path (MEP) for vacancy jump between two adjacent sites.
First, we extract atomistic configurations from MD simulations
at finite temperature, where the vacancy are located at two adjacent sites, as shown in Fig.~\ref{fig:vac}. 
We then perform energy minimization (i.e. relaxation) starting from these configurations to obtain the local energy minima states, as shown in Fig.~\ref{fig:MEP}.
The string relax method~\cite{kuykendall2020stress} is then used to obtain the MEP between these two configurations.
It is important to note that the GNN-MLFF predicts only the force for any atomic configuration but not the potential energy.  Fortunately, the atomic forces are sufficient for the MEP search algorithm (e.g. the string method) to run.  For example, when the algorithm converges, the atomic force should be parallel to the local tangent direction of the MEP everywhere along the path.
After obtaining the MEP, we can numerically integrate the force along the path to obtain the energy barrier, as shown in Fig.~\ref{fig:MEPS}.

The energy barrier obtained from MEP search is $E_{\rm b} = 26.18$~meV for the original LJ potential, and $E_{\rm b} = 25.95$~meV for the GNN-based MLFF (W.O.V.).  
Including the vacancy in the training data does not significantly affect the energy barrier, resulting in $E_{\rm b} = 25.96$~meV for the GNN-based MLFF (W.V.).
Again, it is worth noting that the GNN-based MLFF is able to capture the energy barrier of the vacancy jump even though the corresponding configuration has not been included in the training data.

\section{Conclusions} \label{sec:conclusion}
This paper presents an extensive set of tests that can be used to benchmark the transferability of machine-learned force fields or interatomic potentials for solid-phase simulations. We first demonstrate that the GNN-based MLFF is nearly conservative (i.e. the Hessian matrix is almost symmetric), and we use the Hessian matrix obtained from the MLFF to calculate the PDOS and phonon dispersion relation at 0 K. Furthermore, we compute the phonon dispersion relation using the spectral energy density (SED) method to capture the lattice vibrations at finite temperatures. The phonon dispersion relation and the thermal conductivity from the SED calculation obtained from the GNN-MD (using the MLFF) simulations agree well with the predictions from the original interatomic potential.

Our analysis is then extended to studying defects in the crystal, specifically, the jump rate of a single vacancy. We show that the vacancy jump rate GNN-MD simulations is of the same order of magnitude as the reference rate from the MD simulations. 
This suggests encouraging generalizability of the GNN-based MLFF model because it may be trained on configurations even without a vacancy.
We employed the string relax method to calculate the minimum energy path for vacancy jumps. The resultant energy barrier from the original LJ and the GNN-based MLFF models is in close agreement, providing further evidence for the generalizability of the GNN-based MLFF model.
%

%
\revision{In future work, addressing the challenges posed by the limitations of GNN-based force fields requires a careful analysis of data coverage within the configuration space. This work suggests that the issue lies not in data size but rather in ensuring sufficient diversity to cover a wide range of configurations. For example, the current model explored here is likely to struggle with scenarios such as free surfaces, where the underlying potential energy landscape differs significantly from the training data, and other MLFF architectures may be worth exploring~\cite{poul2023systematic}. Extending this analysis to MLFFs fitted to data generated by more complex potentials, such as the Tersoff potential or EAM potential, could provide additional insights, though we anticipate increased difficulties in accurately capturing many-body effects. Finally, developing robust error quantification metrics depending on physical observables (such as thermal conductivity and vacancy jump rates) beyond forces and energies will be crucial to systematically assess model performance and identify sources of error. Exploring these directions will help guide improvements in the design and training of GNN-based force fields or interatomic potentials.}
The tests presented here demonstrate the promise of MLFFs for studying defects in solids and rare events, which are important for applications in material science. 
%
\revision{Although the GNN-based model benchmarked here is trained on force data from the simple LJ potential, the workflow of training and testing adopted here can be equally applied to MLFFs trained on any force data, including atomic forces obtained from ab initio models.}
These tests can be used to identify the modifications to the dataset that might be required to make the MLFF more generalizable, making high throughput and high-fidelity development of MLFF possible. 
\section*{Conflict of Interest}
The authors declare no competing interests.
\section*{Data Access}
All data can be obtained on request from the corresponding author. The library for the testbed of machine-learned force fields can be found at \href{https://gitlab.com/micronano_public/tb-mlff}{TB-MLFF}. 
\section*{Acknowledgements}
S.M. and W.C. acknowledge support from the Precourt Pioneering Project of Stanford University. S.M. and W.C. acknowledge support from the Air Force Office of Scientific Research under award number FA9550-20-1-0397. We would like to thank Klara Suchan and Myung Chul Kim for their discussions on the manuscript.

\end{document}